%% file: main.tex
\documentclass{ifacconf}

\usepackage{array}
\usepackage{amssymb}
\usepackage{amsmath}
\usepackage{makecell}
\usepackage{transparent}
\usepackage{color}
\usepackage{subcaption}
\usepackage{caption}

\usepackage{import}

\usepackage{graphicx}      
\usepackage{natbib}        
\begin{document}
\begin{frontmatter}

\title{Space-Filling Subset Selection for an Electric Battery Model} 


\author[First]{Philipp Gesner}, 
\author[First]{Christian Gletter}, 
\author[First]{Florian Landenberger}, 
\author[First]{Frank Kirschbaum},
\author[Second]{Lutz Morawietz},
\author[Second]{Bernard Bäker}

\address[First]{Mercedes-Benz AG, Mercedesstr. 137, 70327 Stuttgart}
\address[Second]{Institute of Automotive Technology Dresden, George-Bähr-Str. 1, 01069 Dresden}

\begin{abstract}                
Dynamic models of the battery performance are an essential tool throughout the development process of automotive drive trains. The present study introduces a method making a large data set suitable for modeling the electrical impedance.
When obtaining data-driven models, a usual assumption is that more observations produce better models. However, real driving data on the battery's behavior represent a strongly non-uniform excitation of the system, which negatively affects the modeling. \\
For that reason, a subset selection of the available data was developed. It aims at building accurate \textit{nonlinear autoregressive exogenous} (NARX) models more efficiently. 
The algorithm selects those dynamic data points that fill the input space of the nonlinear model more homogeneously.
It is shown, that this reduction of the training data leads to a higher model quality in comparison to a random subset and a faster training compared to modeling using all data points.
\end{abstract}


\end{frontmatter}

\section{Introduction}
\subsection{Motivation}
The energy storage plays a significant role in the electrification of vehicles. Consequently, characterizing and modeling the performance of lithium-ion batteries is among the higher priorities of many car manufacturers. This paper presents a method to model the nonlinear impedance of automotive batteries based on real driving data. Such data-driven models are especially useful in later development phases.\\   
\cite{LennartLjung.2008} emphasized the role of handling today's extensively collected data for dynamic modeling. However, most publications in the field of \textit{system identification} still rely on a design of input signals. The computational effort and the inhomogeneous excitation are among the reasons preventing a widespread use of large datasets. Accordingly, extracting the relevant information from large datasets has an enormous potential in the context of engineering models.\\
This work aims at obtaining nonlinear models based on large amounts of data from a conventional operation of the battery system. It further tries to reduce the training time while simultaneously generating more accurate models with a space-filling subset selection.
\subsection{Related work}
Modeling the behavior of automotive batteries usually relies on a fundamental understanding of the electrochemical phenomena. Most model structures found in literature capture the basic effects like diffusion and double-layer capacitance by using \textit{equivalent circuit models} (e.g. \cite{Birkl.2013}). Some researchers, for example \cite{Buller.2003}, build even more detailed models integrating additional aspects of the voltage response like the Butler-Volmer relation.\\
Data-driven models have been applied to capture the nonlinear battery dynamics as well, e.g. by \cite{CarlaFabianaChiasserini.2001} using \textit{Markov chains} and by \cite{Capizzi.2011} using \textit{recurrent neural nets}. Both authors demonstrate how black-box models compete with electrochemical models.\\
There are no recognizable efforts to model the battery dynamics based on extensive datasets. In general, studies related to dynamic modeling still lack efficient algorithms for pre-processing large amounts of data. They mostly rely on designed signals like APRBS (\cite{Deflorian.2011}) or chirps (\cite{GokhanAlcan.2018}). Large datasets are almost only utilized for reducing the number of inputs (\cite{SCHMID.2010}) or discovering the underlying system order (\cite{Brunton.2016}.\\
The \textit{mining} of important information within data plays a more significant role in the \textit{machine learning} and \textit{statistics} community. They use well established methods on removing outliers, selecting relevant information and identifying patterns in data (\cite{Han.2011}). \cite{Rennen.2009} for example provides an  overview on data reduction methods for \textit{Kriging} models. He highlights the algorithms, that were developed to solve the \textit{p-dispersion problem} of finding a space-filling subset (\cite{Erkut.1990}).\\
In this context, there are many subset selection methods, e.g. for linear models (\cite{Wang.2019}) or \textit{Bayesian system identification} (\cite{Green.2015}). Recently, \cite{Peter.2019} presented a more holistic approach, which selects points by targeting an arbitrary \textit{probability density function} of the subset. \\ 
The previously discussed algorithms are often exclusive to a particular modeling process or are only validated with relatively small datasets. Hence, the obstacle of this work is an enhancement and combination of those ideas for nonlinear models based on hours of time series data with a high sampling rate.   
\subsection{Example}\label{ch:example}
\begin{figure}[h!]
    \begin{center}
        \def\svgwidth{9.4cm}
        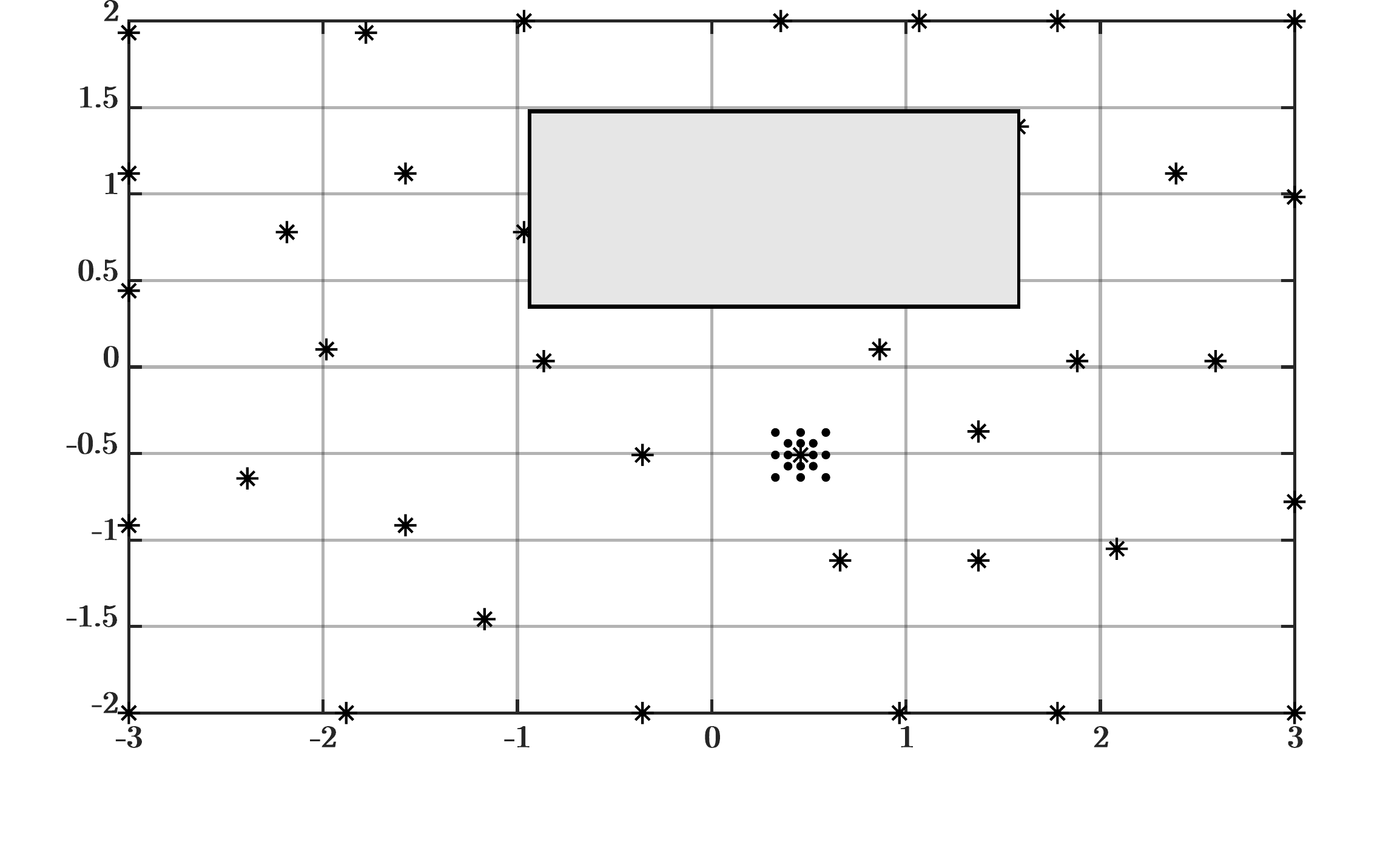    
        \caption{Uniformly distributed data points with additional ones in a certain area} 
        \label{fig:Intro}
    \end{center}
\end{figure}

Modeling a nonlinear function as a black-box is often done with a uniform excitation generated by a space-filling \textit{design of experiment} (DoE). For dynamic systems, the input space additionally includes delayed values of all signals. As a result, most measurements taken during a conventional operation cause a high density close to system's equilibrium.\\
Fig. \ref{fig:Intro} shows an abstracted example of such a scenario. The 49 points represented by the two variables $x_1$ and $x_2$ are designed in a space-filling matter using a \textit{greedy maximin algorithm} (\cite{Steuer.1986}). They are added by 16 points right next to one of the originals.
The variables are part of a \textit{six-hump camel back function} and were used to train a \textit{feedforwad neural net} (FNN) with three layers and 22 neurons. 1000 models were obtained based on the original data points (Model A) and another 1000 models including the added points (Model B).\\
This experiment tries to oppose the common intuition that more data always leads to more accurate models. Using the \textit{Levenberg-Marquart} algorithm and a random initialization for training the FNN, leads to the results in table \ref{tab:example}.
\begin{table}[h!]
    \begin{center}
        \caption{Results based on two different datasets} \label{tab:example}
         \begin{tabular}{lccc}
         	& \parbox[b]{0.1\textwidth}{\centering $\overline{NRMSE}$ \\ (Validation)}  & 
         	\parbox[b]{0.1\textwidth}{\centering $\overline{NRMSE}$ \\ (Training)} & 
         	\parbox[b]{0.1\textwidth}{\centering $\overline{t_T} [s]$} \\ \hline
                Model A & 1.833 & 0.114 & 14.267 \\
                Model B & 2.632 &  0.116 & 17.2 \\ \hline
        \end{tabular}
    \end{center}
\end{table}
\\The outcome, in regard to a broad validation, shows that more data added in a relatively small area of the input space may not only cause longer training times $t_T$, but generate less accurate models. This comparison is based on the same validation dataset and the \textit{normalized root mean squared error} (NRMSE). One explanation for these findings is the equal weighting of all data points during the training. The optimal weights $\omega$ and biases $b$ are determined based on a comparison of the measured output $y(k)$ with the modeled output $\hat{y}(k)$ and can be stated as:
\begin{equation}
\min_{\omega,b}\sum_{k=1}^{n}(y(k)-\hat{y}(k,\omega,b))^2
\label{eq:guetefunktional-ls-summe}
\end{equation}
This summation of all squared errors can lead to an inaccurate representation of the sparser areas and an overweighting of dense areas. As a result, Model A performs better on the validation data.
\section{Battery modeling}
\begin{figure}[h!]
    \centering
    \begin{center}
        \def\svgwidth{8.2cm}
        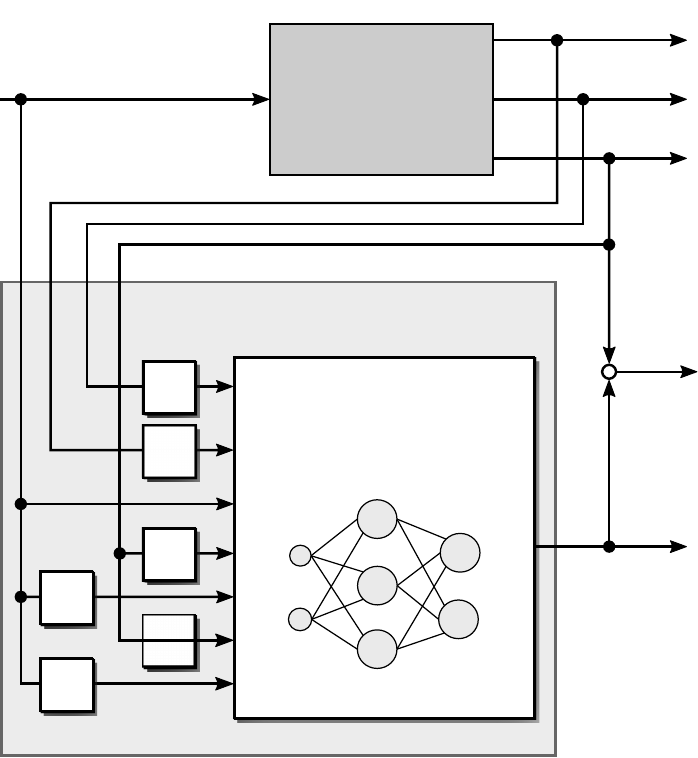    
        \caption{Prediction error of a NARX (\cite{Isermann.2011})}
        \label{fig:NARX}
    \end{center}
\end{figure}

The electric battery dynamics, defined by the chemical cell, consist of many nonlinear processes that can lead to differential equations of high orders. However, the approximation of the voltage response $y$ in one operational point with a 4$^{th}$ order linear system is known to be sufficient for most applications of battery models. Higher order differential equations clearly indicate a pole-zero cancellation (\cite{Scheiffele.2019}). In addition, \cite{Fan.15.12.201518.12.2015} states that a model order reduction  results in an even lower number of remaining states. The common representation with two RC circuits is a special 2$^{nd}$ order model and is used in this study.\\
For building a discrete battery model, it is important  to consider not only the current $u_1$ as input, but also the operational point regarding the temperature $u_2(t)$ and state of charge $u_3(t)$. When it comes to the voltage output, \cite{PhilippGesner.2019} laid out that modeling the battery dynamics of purely electric vehicles up to 5\,Hz is sufficient. \\
Fig. \ref{fig:NARX} illustrates the mentioned signals of the electric battery behavior and defines how the prediction error $e$ of the NARX model is calculated. This focus on the model's one-step prediction in a \textit{series-parallel setup} is used to train the nonlinear regressor. In case of a simulation, the outputs are unknown and the structure is changed to a \textit{parallel setup}, which relies on model outputs which are fed back as new inputs.
The figure further illustrates how every data point of the 7-dimensional input space is mapped to one output. It thereby contains values of shifted signals like $u_1(k-1) = q^{-1}u_1(k)$. Thus, the following nonlinear function of $x(k) =(u_1(k), u_1(k-1), u_1(k-2), u_2(k-1), u_3(k-1), y(k-1), y(k-2))$ needs to be identified:
\begin{equation}\label{eq:FNN}
\hat{y}(k) = \,   f_{FNN}(x(k))
\end{equation}
A \textit{feedforward neural net} (FNN) was chosen for the regression, mainly because of its well established algorithms when it comes to large amounts of observations. It consists of three layers and 22 neurons with \textit{sigmoid activation functions}. Such a small FNN is used to prevent overfitting in this study. For determining an optimum of the cost function (eq. \ref{eq:guetefunktional-ls-summe}) the \textit{Levenberg-Marquart} algorithm is applied. In an attempt to make the modeling more comparable, a stopping criterion was implemented, that depends on a change of the remaining error over the epochs. If its value stays within a band of $10^{-10}$ over $10$ iterations, the training ends.\\
Before the actual modeling an anti-aliasing filter excluding all frequencies above 5\,Hz is applied. The battery signals are then down-sampled to 10\,Hz with a linear interpolation. 
\section{Subset selection}
\begin{figure}[h!]
	\centering
	\begin{subfigure}[l]{0.234\textwidth}
		\caption{All data points}
		\includegraphics[width=\textwidth]{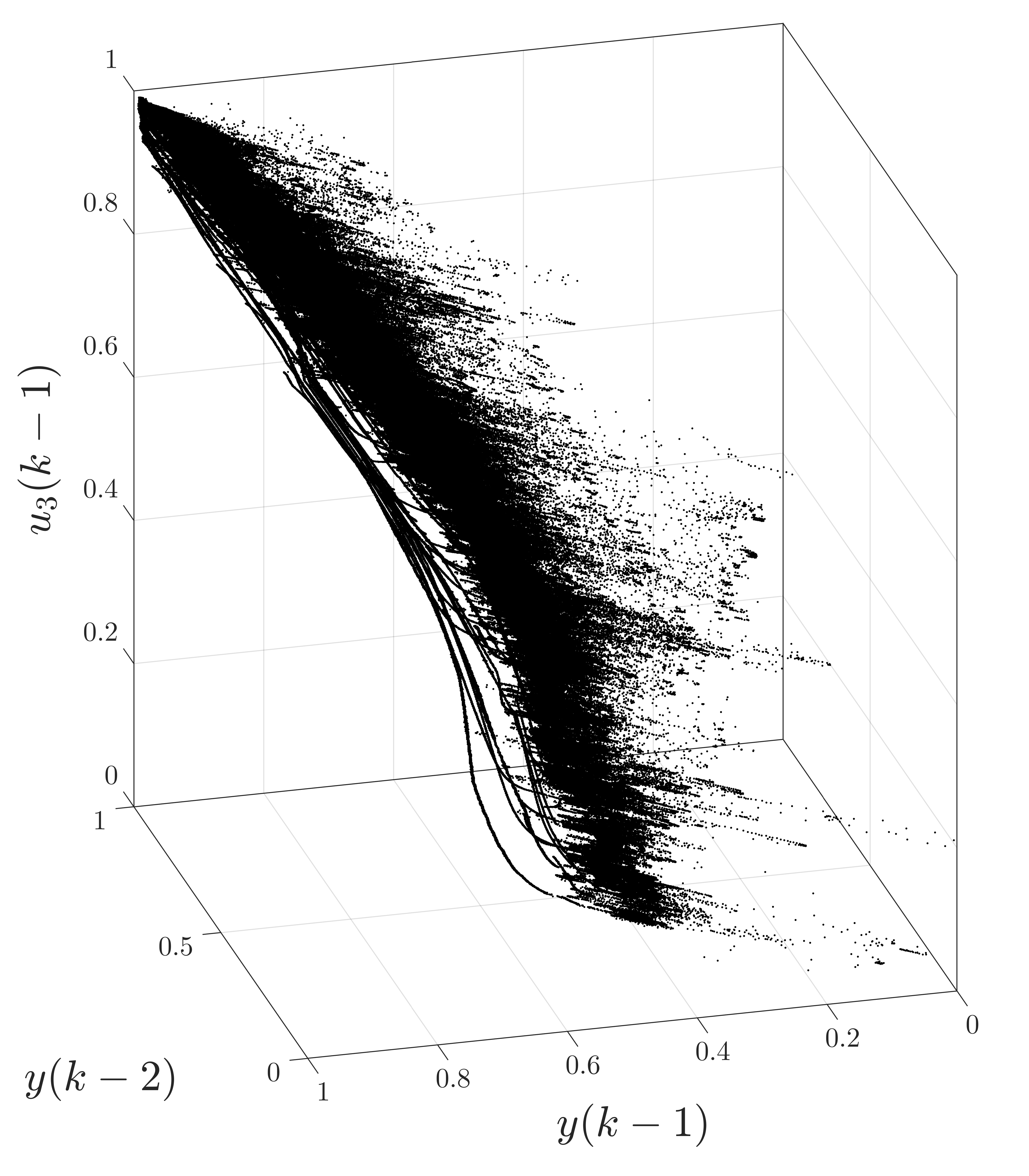} 	
	\end{subfigure}
	~ 
	\begin{subfigure}[l]{0.234\textwidth}	
		\caption{Random subset}
		\includegraphics[width=\textwidth]{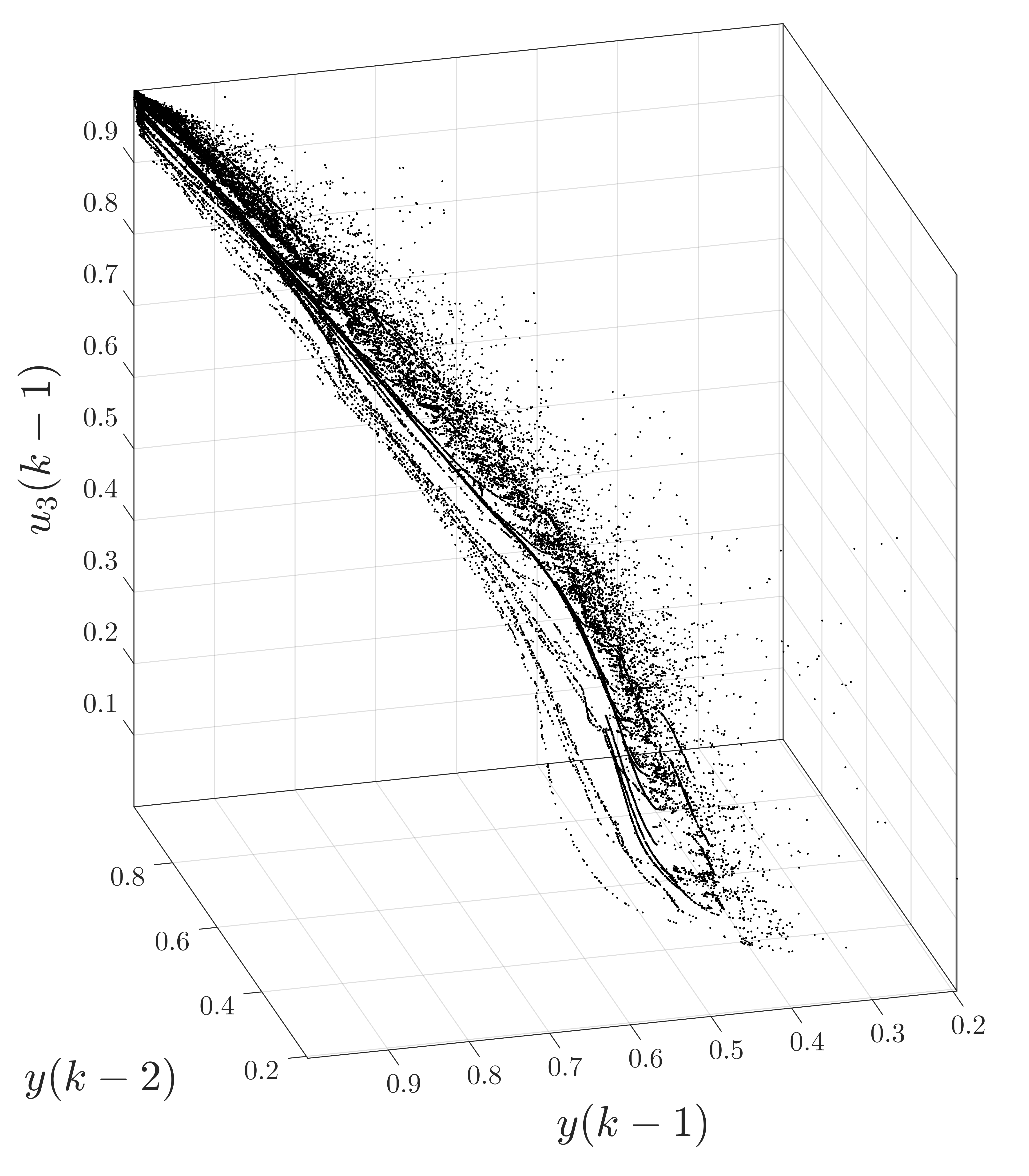} 			
	\end{subfigure}
	~
	\begin{subfigure}[l]{0.234\textwidth}
		\caption{Space-filling subset (LHS)} 
		\includegraphics[width=\textwidth]{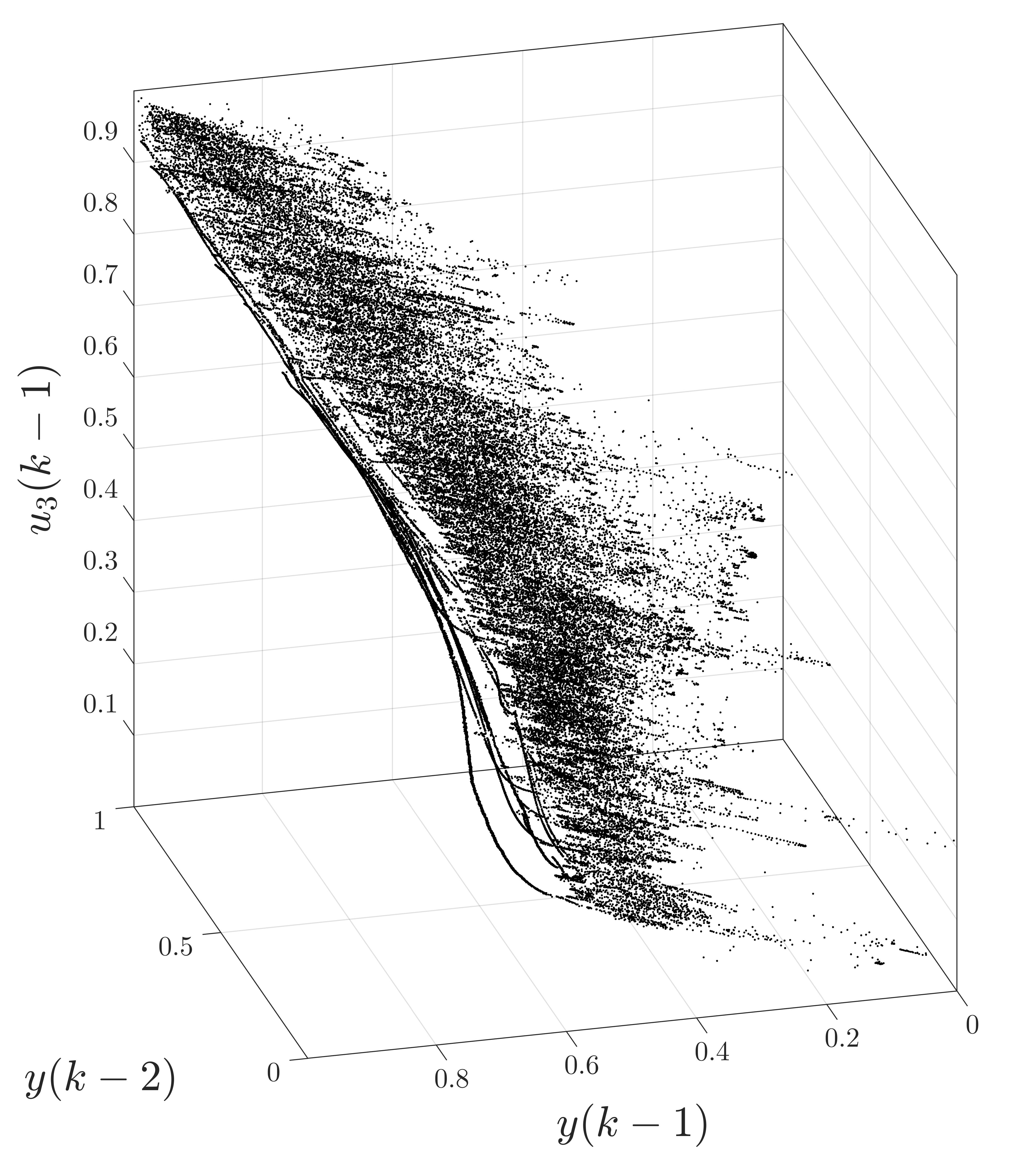} 
	\end{subfigure}
	~ 
	\begin{subfigure}[l]{0.234\textwidth}
		\caption{Space-filling subset (Sobol)} 
		\includegraphics[width=\textwidth]{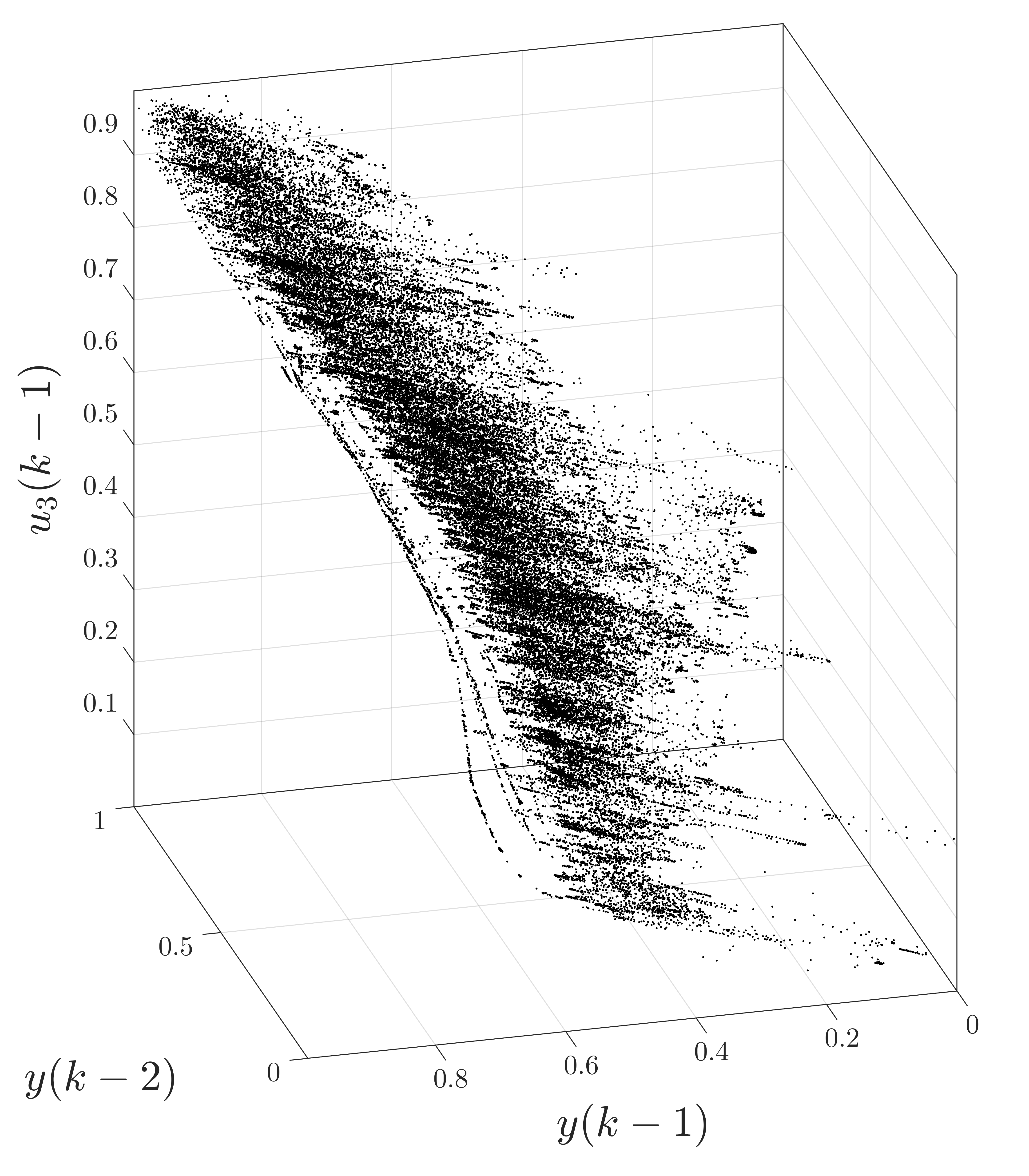} 
	\end{subfigure}
	\caption{Projection onto a subspace of the inputs}\label{fig:Subset}
\end{figure}

The amount of data recorded during the development of a new vehicle is immense. For this study  almost 231\,h of driving data are used, which is only a fraction of the data. Applying the introduced downsampling leads to $n_{all}$ = 8,312,122 discrete time steps for the training.\\
As it is natural for real driving, most of the available battery data is concentrated near its equilibrium. The introductory example in mind (chapter \ref{ch:example}), those dense areas can lead to unnecessary long training times and globally inaccurate models. Hence, there is a need of reducing the redundancy.\\
In the context of nonlinear data models, a space-filling design of the inputs is a regular choice. The idea is to distribute points as uniformly as possible in any dimension $d$. The following \textit{measure of coverage} $\lambda$ is one of the adequate metrics for this criteria:
\begin{equation} \label{CM}
\lambda = \dfrac{1}{\overline{v}}\sqrt{\frac{1}{n}\sum_{i=1}^{n}{(v_i - \overline{v})^2}}
\end{equation}
It compares the minimal distance $v_i =  min \, {\|{x_{i} - x}\|_2}$ between each point $x_i$ and all other points $x$ with the mean of those distances $\overline{v}$. The smaller $\lambda$ is, the more uniform becomes the point distribution. Consequently, the metric is at the core of many algorithms for a space-filling subset selection. \\
Considering the 7-dimensional input space (Fig. \ref{eq:FNN}) and the large number of observations, finding a homogeneous subset based on $\lambda$ is computationally intensive. For example the \textit{greedy maxmin algorithm} applied in chapter \ref{ch:example} already indicates an exponential time complexity concerning the number of samples.\\
In this study, an approach is suggested that designs a space-filling experiment regardless of the recorded data. Afterwards, the data points closest to the designed points $x_{DoE}$ are selected. The following generalized steps result in such a subset containing $n_{sub} = \alpha \, n_{all}$ points of the original data:
\begin{enumerate}
	\item Normalize $n_{all}$ given data points: $x_{N} \in 	(\mathbb{R}^{d} \cap [0,\: 1])$
	\item Design $n_{DoE}(\alpha) =  \alpha \, \frac{n_{all}}{V_{all}} $ uniformly distributed points in $\mathbb{R}^{d} \cap [0, 1]$\label{DOE}
	\item Pick their \textit{nearest neighbour}: $min({\|{x_{DoE,i} - x_{N}}\|_2})$
\end{enumerate}
Volume $V_{all}$, spanned by the measured points in $\mathbb{R}^{7}$, is approximated by a convex hull. For the space-filling design two popular methods were chosen. The \textit{Sobol Sequence} is used because of its simplistic calculation and the \textit{Latin Hypercube Sampling} (LHS) because of the ability to generate a point distribution that is still space-filling in projections onto its subspaces. For validation reasons, they are added by a randomly selected subset. Fig. \ref{fig:Subset} shows a projection of the data and its subsets onto $u_3(k)$, $y(k-1)$ and $y(k-2)$.
\\The LHS and \textit{Sobol} subset, containing 1\,\% of all data points, achieve a similar space-fillingness of $\lambda \approx 0.01$. In comparison to the random subsets and the original dataset with $\lambda \approx 3.5$, the algorithm undoubtedly leads to more homogeneous data distribution. Computing such a subset (\textit{Sobol}) takes about 18 minutes on a Intel(R) Core(TM) i7-7820HQ CPU with a clock rate of 2.90\,GHz.
\section{Results and conclusion}
\begin{figure}[h!]
    \begin{subfigure}[l]{0.49\textwidth}
    	\caption{Driving data with a pdf similar to the training}
        \includegraphics[width=\textwidth]{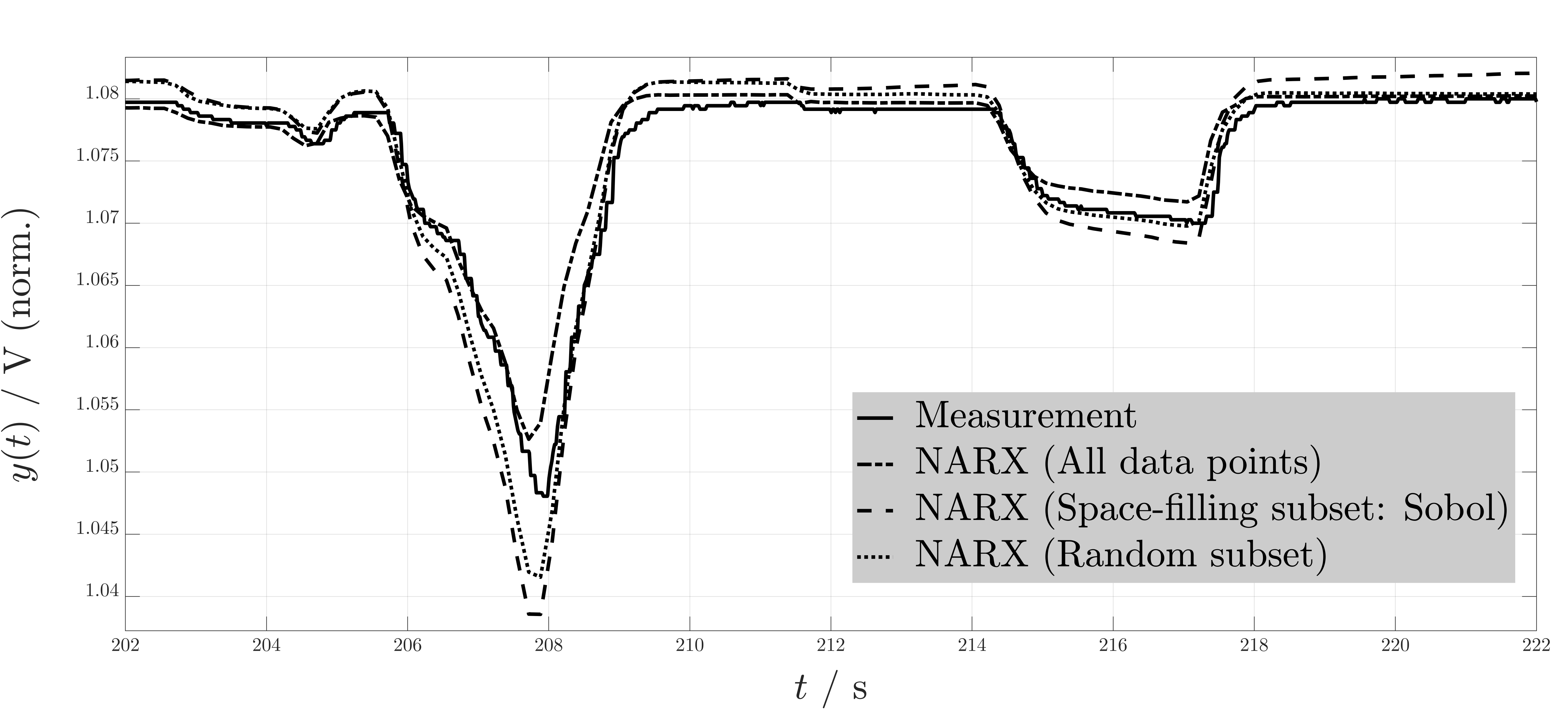}
        \label{fig:validation1}
    \end{subfigure}
    \begin{subfigure}[l]{0.49\textwidth}
    	\caption{Driving data with a strongly distinguishable pdf from the training}
        \includegraphics[width=\textwidth]{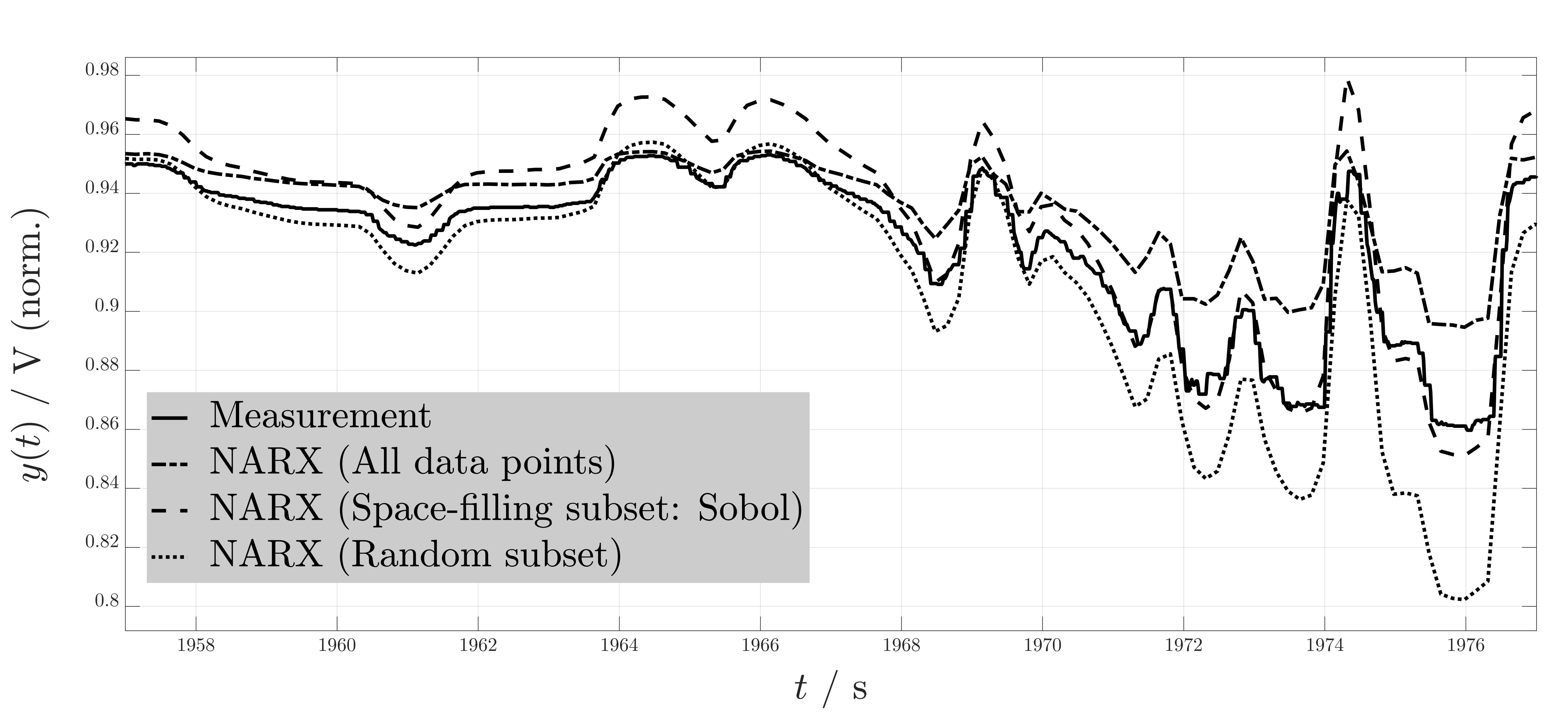}
        \label{fig:validation2}
    \end{subfigure}
    \caption{Comparing simulation performance of the models} \label{fig:Validation}
\end{figure}

To validate the presented space-filling subset selection, two additional recordings were picked. One of them matches the \textit{probability density function} (pdf) of the training data quite well (Validation 1). The other one (Validation 2) is chosen to test the models capability of simulating the battery's behavior globally. Its data distribution in the input space is strongly distinguishable from the training data set. This comparison was conducted with an estimate of the pdf using the MATLAB function \textit{mvksdensity}.
\\ Fig. \ref{fig:Validation} shows those results of simulating the electric battery behavior. For the space-filling approach, only the model trained on the \textit{Sobol} subset is shown because of illustrative reasons.
\\The graphs already indicate that the space-filling subset outperforms the random subset regarding validation datasets. The visual error between the model trained on all data and the LHS subset is rather small. Note, that the plotted parts of the driving data do not reflect the entire performance of the models. Therefore, table \ref{tb:margins} states more precise results, based on three models trained per dataset.
\begin{table}[h!]
    \begin{center}
        \caption{Validation results of the model's performance}\label{tb:margins}
        \begin{tabular}{lccc}
        	            & \parbox[b]{0.1\textwidth}{\centering $\overline{NRMSE}$ \\ (Validation\medspace 1)}   &  \parbox[b]{0.1\textwidth}{\centering $\overline{NRMSE}$ \\ (Validation\medspace 2)} & \parbox[b]{0.1\textwidth}{\centering $\overline{t_T} [h]$}  \\ \hline
            All data points  & 0.49 &  2.43 &  11.63  \\
           	Sobol subset & 0.52 &  1.86 &  1.96  \\ 
            LHS subset  & 0.67 &  2.3 &  1.07  \\
            Random subset & 1.08 &  5.9 &  0.38\\ \hline
        \end{tabular}
    \end{center}
\end{table}
The advantage of subset selection in terms of training time $\overline{t_T}$ is undeniable. Additionally, the NRMSE, normalized by the nominal battery voltage, was used toquantify the model's performance. Both space-filling subsets result in more accurate models than the random subset. Concerning the second dataset, which represents a more unconventional operation of the battery, a space-filling subset leads to better models than using all data points.\\
In conclusion, overweighting some parts of the input space can cause inaccuracies of the resulting models. This calls for more research on pre-selecting data in the field of dynamic system identification. 
\bibliography{IFAC_WC}             
\end{document}

%% file: Example.pdf_tex
\begingroup%
  \makeatletter%
  \providecommand\color[2][]{%
    \errmessage{(Inkscape) Color is used for the text in Inkscape, but the package 'color.sty' is not loaded}%
    \renewcommand\color[2][]{}%
  }%
  \providecommand\transparent[1]{%
    \errmessage{(Inkscape) Transparency is used (non-zero) for the text in Inkscape, but the package 'transparent.sty' is not loaded}%
    \renewcommand\transparent[1]{}%
  }%
  \providecommand\rotatebox[2]{#2}%
  \newcommand*\fsize{\dimexpr\f@size pt\relax}%
  \newcommand*\lineheight[1]{\fontsize{\fsize}{#1\fsize}\selectfont}%
  \ifx\svgwidth\undefined%
    \setlength{\unitlength}{664.42277478bp}%
    \ifx\svgscale\undefined%
      \relax%
    \else%
      \setlength{\unitlength}{\unitlength * \real{\svgscale}}%
    \fi%
  \else%
    \setlength{\unitlength}{\svgwidth}%
  \fi%
  \global\let\svgwidth\undefined%
  \global\let\svgscale\undefined%
  \makeatother%
  \begin{picture}(1,0.60121656)%
    \lineheight{1}%
    \setlength\tabcolsep{0pt}%
    \put(0,0){\includegraphics[width=\unitlength,page=1]{Example.pdf}}%
    \put(0.44677923,0.47465044){\color[rgb]{0,0,0}\makebox(0,0)[lt]{\lineheight{1.25}\smash{\begin{tabular}[t]{l}Original points\end{tabular}}}}%
    \put(0.44781827,0.39906001){\color[rgb]{0,0,0}\makebox(0,0)[lt]{\lineheight{1.25}\smash{\begin{tabular}[t]{l}Added points\end{tabular}}}}%
    \put(0.49350685,0.00864237){\color[rgb]{0,0,0}\makebox(0,0)[lt]{\lineheight{1.25}\smash{\begin{tabular}[t]{l}$x_{1}$\end{tabular}}}}%
    \put(1,0.25726367){\color[rgb]{0,0,0}\rotatebox{90}{\makebox(0,0)[lt]{\lineheight{1.25}\smash{\begin{tabular}[t]{l}  \end{tabular}}}}}%
    \put(0,0){\includegraphics[width=\unitlength,page=2]{Example.pdf}}%
    \put(-0.00091053,0.3248018){\color[rgb]{0,0,0}\makebox(0,0)[lt]{\lineheight{1.25}\smash{\begin{tabular}[t]{l}$x_{2}$\end{tabular}}}}%
  \end{picture}%
\endgroup%

%% file: NARX_Prediction.pdf_tex
\begingroup%
  \makeatletter%
  \providecommand\color[2][]{%
    \errmessage{(Inkscape) Color is used for the text in Inkscape, but the package 'color.sty' is not loaded}%
    \renewcommand\color[2][]{}%
  }%
  \providecommand\transparent[1]{%
    \errmessage{(Inkscape) Transparency is used (non-zero) for the text in Inkscape, but the package 'transparent.sty' is not loaded}%
    \renewcommand\transparent[1]{}%
  }%
  \providecommand\rotatebox[2]{#2}%
  \newcommand*\fsize{\dimexpr\f@size pt\relax}%
  \newcommand*\lineheight[1]{\fontsize{\fsize}{#1\fsize}\selectfont}%
  \ifx\svgwidth\undefined%
    \setlength{\unitlength}{200.819538bp}%
    \ifx\svgscale\undefined%
      \relax%
    \else%
      \setlength{\unitlength}{\unitlength * \real{\svgscale}}%
    \fi%
  \else%
    \setlength{\unitlength}{\svgwidth}%
  \fi%
  \global\let\svgwidth\undefined%
  \global\let\svgscale\undefined%
  \makeatother%
  \begin{picture}(1,1.08569652)%
    \lineheight{1}%
    \setlength\tabcolsep{0pt}%
    \put(0,0){\includegraphics[width=\unitlength,page=1]{NARX_Prediction.pdf}}%
    \put(0.46881373,0.9316883){\color[rgb]{0,0,0}\makebox(0,0)[lt]{\lineheight{1.25}\smash{\begin{tabular}[t]{l}\textbf{Battery}\end{tabular}}}}%
    \put(0.42716627,0.48997351){\color[rgb]{0,0,0}\transparent{0.95999998}\makebox(0,0)[lt]{\lineheight{1.25}\smash{\begin{tabular}[t]{l}Feed Forward\end{tabular}}}}%
    \put(0.45380902,0.44171162){\color[rgb]{0,0,0}\transparent{0.95999998}\makebox(0,0)[lt]{\lineheight{1.25}\smash{\begin{tabular}[t]{l}Neural Net\end{tabular}}}}%
    \put(0.89243401,0.57581147){\color[rgb]{0,0,0}\makebox(0,0)[lt]{\lineheight{1.25}\smash{\begin{tabular}[t]{l}$e(k)$\end{tabular}}}}%
    \put(0.22810222,0.63147179){\color[rgb]{0,0,0}\transparent{0.95999998}\makebox(0,0)[lt]{\lineheight{1.25}\smash{\begin{tabular}[t]{l}\textbf{NARX Model}\end{tabular}}}}%
    \put(0.81071881,0.58503936){\color[rgb]{0,0,0}\transparent{0.95999998}\makebox(0,0)[lt]{\lineheight{1.25}\smash{\begin{tabular}[t]{l}$+$\end{tabular}}}}%
    \put(0.81071881,0.49419245){\color[rgb]{0,0,0}\transparent{0.95999998}\makebox(0,0)[lt]{\lineheight{1.25}\smash{\begin{tabular}[t]{l}$-$\end{tabular}}}}%
    \put(0.89279478,0.87946229){\color[rgb]{0,0,0}\makebox(0,0)[lt]{\lineheight{1.25}\smash{\begin{tabular}[t]{l}$y(k)$\end{tabular}}}}%
    \put(0.89416127,0.96248728){\color[rgb]{0,0,0}\makebox(0,0)[lt]{\lineheight{1.25}\smash{\begin{tabular}[t]{l}$u_3(k)$\end{tabular}}}}%
    \put(0.89419614,1.0473774){\color[rgb]{0,0,0}\makebox(0,0)[lt]{\lineheight{1.25}\smash{\begin{tabular}[t]{l}$u_2(k)$\end{tabular}}}}%
    \put(0.89469431,0.32360535){\color[rgb]{0,0,0}\makebox(0,0)[lt]{\lineheight{1.25}\smash{\begin{tabular}[t]{l}$\hat{y}(k)$\end{tabular}}}}%
    \put(0.0336014,0.96993396){\color[rgb]{0,0,0}\makebox(0,0)[lt]{\lineheight{1.25}\smash{\begin{tabular}[t]{l}$u_{1}(k)$\end{tabular}}}}%
    \put(0.06419085,0.21404607){\color[rgb]{0,0,0}\makebox(0,0)[lt]{\lineheight{1.25}\smash{\begin{tabular}[t]{l}$q^{-1}$\end{tabular}}}}%
    \put(0.21206308,0.27629778){\color[rgb]{0,0,0}\makebox(0,0)[lt]{\lineheight{1.25}\smash{\begin{tabular}[t]{l}$q^{-1}$\end{tabular}}}}%
    \put(0.21162607,0.42456729){\color[rgb]{0,0,0}\makebox(0,0)[lt]{\lineheight{1.25}\smash{\begin{tabular}[t]{l}$q^{-1}$\end{tabular}}}}%
    \put(0.21206308,0.51583722){\color[rgb]{0,0,0}\makebox(0,0)[lt]{\lineheight{1.25}\smash{\begin{tabular}[t]{l}$q^{-1}$\end{tabular}}}}%
    \put(0,0){\includegraphics[width=\unitlength,page=2]{NARX_Prediction.pdf}}%
    \put(0.21206308,0.15174783){\color[rgb]{0,0,0}\makebox(0,0)[lt]{\lineheight{1.25}\smash{\begin{tabular}[t]{l}$q^{-2}$\end{tabular}}}}%
    \put(0.06419085,0.08944378){\color[rgb]{0,0,0}\makebox(0,0)[lt]{\lineheight{1.25}\smash{\begin{tabular}[t]{l}$q^{-2}$\end{tabular}}}}%
  \end{picture}%
\endgroup%

%% file: main.bbl
\begin{thebibliography}{20}
\providecommand{\natexlab}[1]{#1}
\providecommand{\url}[1]{\texttt{#1}}
\providecommand{\urlprefix}{URL }
\expandafter\ifx\csname urlstyle\endcsname\relax
  \providecommand{\doi}[1]{doi:\discretionary{}{}{}#1}\else
  \providecommand{\doi}{doi:\discretionary{}{}{}\begingroup
  \urlstyle{rm}\Url}\fi

\bibitem[{Birkl(2013)}]{Birkl.2013}
Birkl, C.R. (2013).
\newblock {Model identification and parameter estimation for LiFePO4
  batteries}.
\newblock In \emph{{IET Hybrid and Electric Vehicles Conference}}. IEEE,
  Piscataway, NJ.

\bibitem[{Brunton(2016)}]{Brunton.2016}
Brunton, S.L. (2016).
\newblock {Discovering governing equations from data by sparse identification
  of nonlinear dynamical systems}.
\newblock \emph{{Proceedings of the National Academy of Sciences of the United
  States of America}}, 113.

\bibitem[{Buller(2003)}]{Buller.2003}
Buller, S. (2003).
\newblock \emph{{Impedance-based simulation models for energy storage devices
  in advanced automotive power systems}}.
\newblock {Aachener Beitr{\"a}ge des ISEA}. Shaker, Aachen.

\bibitem[{Capizzi(2011)}]{Capizzi.2011}
Capizzi, G. (2011).
\newblock {Recurrent Neural Network-Based Modeling and Simulation of Lead-Acid
  Batteries Charge--Discharge}.
\newblock \emph{{IEEE Transactions on Energy Conversion}}, 26.

\bibitem[{Chiasserini(2001)}]{CarlaFabianaChiasserini.2001}
Chiasserini, C.F. (2001).
\newblock {Energy efficient battery management}.
\newblock \emph{{IEEE Journal on selected areas in communications}}, 19.

\bibitem[{Deflorian(2011)}]{Deflorian.2011}
Deflorian, M. (2011).
\newblock {Design of Experiments for nonlinear dynamic system identification}.
\newblock \emph{{IFAC Proceedings Volumes}}, 44.

\bibitem[{Erkut(1990)}]{Erkut.1990}
Erkut, E. (1990).
\newblock {The discrete p-dispersion problem}.
\newblock \emph{{European Journal of Operational Research}}, 46(1).

\bibitem[{Fan(2015)}]{Fan.15.12.201518.12.2015}
Fan, G. (2015).
\newblock {A comparison of model order reduction techniques for electrochemical
  characterization of Lithium-ion batteries}.
\newblock In IEEE (ed.), \emph{{IEEE Conference on Decision and Control
  (CDC)}}.

\bibitem[{Gesner(2019)}]{PhilippGesner.2019}
Gesner, P. (2019).
\newblock {Modeling and identification of electrochmical enery storage for
  drive train development}.
\newblock In {VDI Verlag} (ed.), \emph{{19th International Congress ELIV}}.

\bibitem[{Gokhan(2018)}]{GokhanAlcan.2018}
Gokhan, A. (2018).
\newblock {Diesel Engine NOx Emission Modeling Using a New Experiment Design
  and Reduced Set of Regressors}.
\newblock \emph{{IFAC-PapersOnLine}}, 51.

\bibitem[{Green(2015)}]{Green.2015}
Green, P.L. (2015).
\newblock {Bayesian system identification of dynamical systems using large sets
  of training data: A MCMC solution}.
\newblock \emph{{Probabilistic Engineering Mechanics}}, 42.

\bibitem[{Han(2011)}]{Han.2011}
Han, J. (2011).
\newblock \emph{{Data Mining: Concepts and Techniques: Concepts and
  Techniques}}.
\newblock {Elsevier professional}.

\bibitem[{Isermann(2011)}]{Isermann.2011}
Isermann, R. (2011).
\newblock \emph{{Identification of Dynamic Systems}}.
\newblock Springer, Berlin.

\bibitem[{Ljung(2008)}]{LennartLjung.2008}
Ljung, L. (2008).
\newblock {Perspectives on System Identification}.
\newblock \emph{{IFAC Proceedings Volumes}}, 41.

\bibitem[{Peter and Nelles(2019)}]{Peter.2019}
Peter, T.J. and Nelles, O. (2019).
\newblock {Fast and simple dataset selection for machine learning}.
\newblock \emph{{at - Automatisierungstechnik}}, (10).

\bibitem[{Rennen(2009)}]{Rennen.2009}
Rennen, G. (2009).
\newblock {Subset selection from large datasets for Kriging modeling}.
\newblock \emph{{Structural and Multidisciplinary Optimization}}, 38(6).

\bibitem[{Scheiffele(2019)}]{Scheiffele.2019}
Scheiffele, J. (2019).
\newblock \emph{{Parameter estimation methods for models of current and voltage
  behavior in lithium-ion batteries}}.
\newblock {Master Thesis}, {University of Stuttgart}.

\bibitem[{Schmid(2010)}]{SCHMID.2010}
Schmid, P. (2010).
\newblock {Dynamic mode decomposition of numerical and experimental data}.
\newblock \emph{{J. Fluid Mech.}}, 656.

\bibitem[{Steuer(1986)}]{Steuer.1986}
Steuer, R.E. (1986).
\newblock \emph{{Multiple criteria optimization: Theory, computation, and
  application}}.
\newblock {Wiley series in probability and mathematical statistics}.

\bibitem[{Wang(2019)}]{Wang.2019}
Wang, H. (2019).
\newblock {Divide-and-Conquer Information-Based Optimal Subdata Selection
  Algorithm}.
\newblock \emph{{J Stat Theory Pract}}, 13(3).

\end{thebibliography}
